\documentclass[letterpaper]{article} 
\usepackage{aaai24}
\usepackage{times}  
\usepackage{helvet}  
\usepackage{courier}  
\usepackage[hyphens]{url}  
\usepackage{graphicx} 
\usepackage{natbib}  
\usepackage{caption} 
\usepackage{newfloat}
\usepackage{listings}
\DeclareCaptionStyle{ruled}{labelfont=normalfont,labelsep=colon,strut=off} 
\usepackage{amsfonts}       
\usepackage{amsmath}
\usepackage{amsthm}
\usepackage{amssymb}
\usepackage{subcaption}
\usepackage[linesnumbered,ruled, vlined]{algorithm2e}
\usepackage{multirow}

\title{Ternary Singular Value Decomposition as a Better Parameterized Form in Linear Mapping}

\author{%
    \begin{tabular}{c c c }
    	Boyu Chen & Hanxuan Chen & Jiao He\\
    	\texttt{chenboyu17@huawei.com} & \texttt{chenhanxuan@hisilicon.com} & \texttt{hejiao4@huawei.com}
    \end{tabular}
    \begin{tabular}{c c}
    	 Fengyu Sun & Shangling Jui\\\texttt{sunfengyu@huawei.com} & \texttt{jui.shangling@huawei.com}
    \end{tabular}
}
\nocopyright

\DeclareMathOperator{\diag}{diag}
\newtheorem{theorem}{Theorem}
\newtheorem{remark}{remark}
\newtheorem{corollary}{Corollary}
\newtheorem{lemma}{Lemma}

\begin{document}

\maketitle

\begin{abstract}
  We present a simple yet novel parameterized form of linear mapping to achieves remarkable network compression performance: a pseudo SVD called Ternary SVD (TSVD). 
  Unlike vanilla SVD, TSVD limits the $U$ and $V$ matrices in SVD to ternary matrices form in $\{\pm 1, 0\}$. This means that instead of using the expensive multiplication instructions, TSVD only requires addition instructions when computing $U(\cdot)$ and $V(\cdot)$.
  We provide direct and training transition algorithms for TSVD like Post Training Quantization and Quantization Aware Training respectively. Additionally, we analyze the convergence of the direct transition algorithms in theory. 
  In experiments, we demonstrate that TSVD can achieve state-of-the-art network compression performance in various types of networks and tasks, including current baseline models such as ConvNext, Swim, BERT, and large language model like OPT.
\end{abstract}

\section{Introduction}

Linear mapping, which includes fully connected layers and convolution layers, is a crucial component of modern neural networks in most cases. It accounts for virtually all parameter counts and FLOPS of the entire network and is always the primary target in network compression. 

The current weight compression method for a linear mapping can be broadly classified into three principles \cite{neill2020overview}: quantization, low rank decomposition, and pruning. Many current works, including \cite{zafrir2021prune, li2021towards, guo2022compact, jaderberg2014speeding, frantar2023sparsegpt, frantar2022gptq},  focus solely on improving the fine-tuning and calibration procedures based on these principles. However, the representation accuracy of these principles limits the upper limit of their effectiveness, particularly in the case of super low-bit quantization.

In recent years, a new principle has emerged: using cheap addition instructions instead of the expensive multiplication instructions for acceleration. This approach has been explored in various works, including \cite{chen2020addernet,you2020shiftaddnet,courbariaux2015binaryconnect}. However, all of these approaches require building a specific model architecture and training from scratch, which makes them impractical for large language models.

In this paper, based on this new principle, we propose Ternary SVD (TSVD) as an improved parameterized form of linear mapping, building upon the ideas of SVD and AdderNet\cite{chen2020addernet}. TSVD limits the $U$ and $V$ matrices of SVD into two Ternary matrices in $\{\pm 1, 0\}$. Unlike vanilla SVD, the rank of TSVD is typically not small. Based on the results of fitting random matrices in Figure \ref{fig:TSVD_profile}, it is evident that our TSVD approach, which is based on the new principle, outperforms quantization, low rank decomposition and pruning. In short, our contributions in this paper are as follows:
\begin{itemize}
    \item We introduce TSVD as a new parameterized form of linear mapping, which is significantly accelerated by replacing multiplication instructions with sparsity additions. To the best of our knowledge, TSVD is the first data-independent ternary PTQ method that is suitable for a wide range of network scales and tasks while maintaining high accuracy.
    \item We analyze the convergence of the direct transition algorithm in theory. Also, we introduce a simple yet novel way of STE in TSVD QAT algorithms.
    \item In experiments, we demonstrate that TSVD can achieve state-of-the-art network compression performance in various types of networks and tasks, including current baseline models such as ConvNext, Swim, BERT, and large language model like OPT.
\end{itemize}

\section{Preliminary}

\subsection{Truncated SVD in Network Compression}

We will begin by examining the compression of vanilla SVD in a fully connected layer. Consider a scenario where there is a single input sample on the fully connected layer without any bias. The equation is as follows: 
\begin{equation}\label{eq:fc}
    y=Wx
\end{equation}
where $W$ is a parameter matrix with shape $[M, N]$, $y$ is the output vector with shape $[M, 1]$, and $x$ is an input vector with shape $[N, 1]$. Vanilla SVD decomposes $W$ into three parameter tensors $U, S,$ and $V$ by solving the following optimization problem: 
\begin{equation}\label{eq:opt_svd}
U, S, V = \mathop{\arg\min}_{U, S, V} \|U\diag(S)V - W\|_F  
\end{equation}
where $U$ is a column orthogonal matrix of shape $[M, K]$, $V$ is a row orthogonal matrix of shape $[K, N]$, and $S$ is a singular vector of shape $[K]$. If $U \diag(S)V$ can closely approximate $W$, then eq \ref{eq:fc} can be computed using the following formula: 
\[
  y=Wx \simeq U(\diag(S)(Vx))
\]
For vanilla SVD, a low rank $K$ is required for network compression. However, in practice, achieving a good approximation of $W$ under such low $K$ conditions can be challenging. Previous SVD methods for network compression have often faced difficulties in balancing the tradeoff between approximation and acceleration. The critical rank $K$ (denoted as $\bar{K}$) that balances FLOPS with and without vanilla SVD is simply:
\[
  \bar{K} = \frac{M N}{ M + N}
\]

\subsection{Hardware Cost of Basic Instruction}

For hardware implementation, the addition instruction is often much cheaper than multiplication. Previous works have proven effective by making the most use of feature, including \cite{courbariaux2015binaryconnect,lavin2016fast,chen2020addernet,you2020shiftaddnet}. However, accurately comparing their costs depends on the specific hardware and which cost we are concerned with. Theoretical estimates suggest that $d$-bit addition costs $\mathcal{O}(d)$ and multiplication costs $\mathcal{O}(d^2)$, which aligns with the energy data from \cite{you2020shiftaddnet,zhang2022pokebnn}. However, estimating latency cost in such way is usually too ideal, although the order of magnitude is still correct. In this paper, since there always exist a sign bit in $d$-bit presentation and there is $d-2$ addition in the long multiplication algorithm of $d-1$ bit unsigned integer, to avoid overcomplicating the problem, we simply assume that 
\[1 \times \text{Mul} = (d - 2) \times \text{Add} \]
for principle elucidation and experiment intuitive understanding, while also providing actual multiplication and addition counts in our experiments for custom assessment. 

\section{Ternary SVD}
\begin{figure*}[tb]
  \centering
  \begin{subfigure}{.49\textwidth}
    \centering
    \includegraphics[width=\linewidth]{"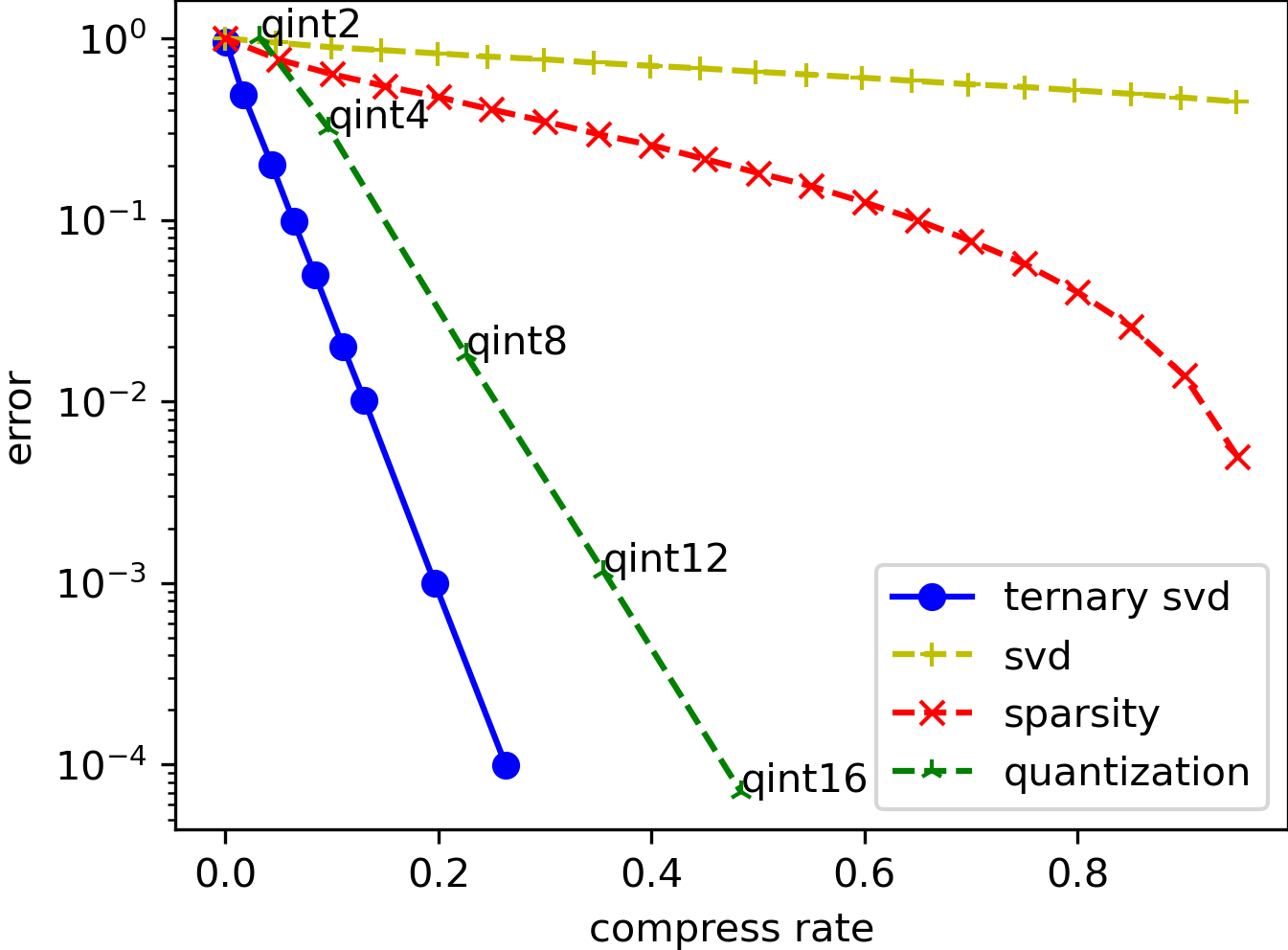"}
    \caption{}\label{fig:TSVD_profile}
  \end{subfigure}
  \begin{subfigure}{.44\textwidth}
    \centering
    \includegraphics[width=\linewidth]{"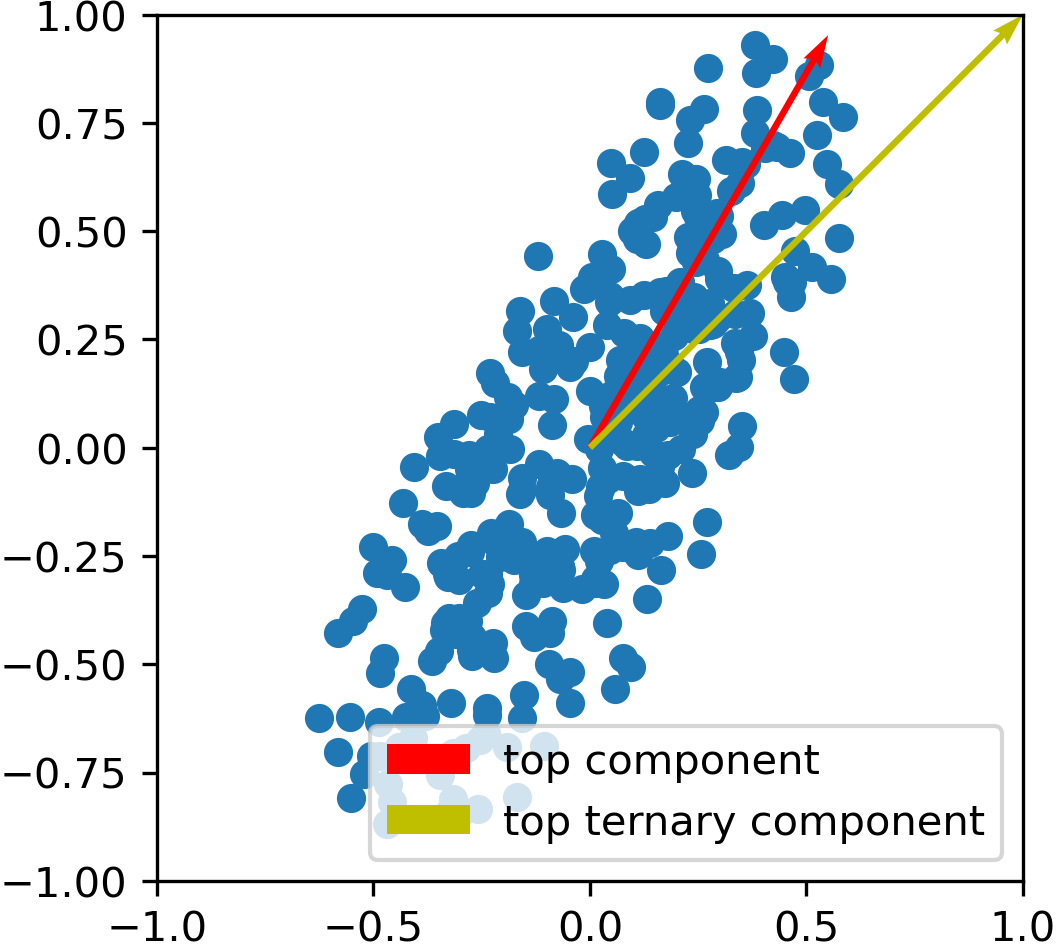"}
    \caption{}\label{fig:greedy}
  \end{subfigure}
  \caption{(a): For various network weight compression methods, tradeoff between compression rate, and relative operator norm error on a $[512, 256]$ float32 random matrix with Laplace distribution. Our method is represented by the blue line. For more detail, see section \ref{sec:TSVD_profile}. (b): Illustration of the main belief of our greedy algorithm \ref{alg:binary_svd} in 2 dimension. We can view the SVD decomposition as a PCA process to the column or row vector of $W$. The goal of TSVD definition is to find the top ternary component (indicated by the yellow arrow). To achieve this, we first identify the top component (indicated by the red arrow) and then convert it into its closest ternarization vector.}
\end{figure*}

Ternary SVD, regardless of how to solve it, can also be defined as the optimization problem of \eqref{eq:opt_svd}, with the additional constraint that $U$ and $V$ are both Ternary matrices in $\{\pm 1, 0\}$. This allows for the use of addition instructions instead of multiplication instructions in computing $U(\cdot)$ and $V(\cdot)$, resulting in faster computation. Compared to vanilla SVD, TSVD has a more relaxed constraint on the rank $K$ for acceleration. The critical rank $\bar{K}$ of Ternary SVD is given by
\begin{equation}\label{eq:tsvd_critical_rank}
  \bar{K} = \frac{(d-1) M N}{d + r(M + N) - 2}
\end{equation}
where $d$ is the bit-width and $r$ is the sparsity rate (non-zero rate) of $U$ and $V$ (a detailed derivation can be found in section \ref{sec:tsvd_critical_rank_detail}). As shown in figure \ref{fig:TSVD_profile}, it is evident that TSVD outperforms previous compression principles in terms of the tradeoff between compression rate and approximation error in a simple random matrix test with Laplace distribution.

\subsection{Direct Transition Algorithm to TSVD form}
We have developed a simple greedy algorithm to perform this transformation, which is based on the main belief that \emph{the largest TSVD component of $W$ should be close to the simply ternarization of the largest SVD component} (illustrated in figure \ref{fig:greedy}). We begin by initializing the residual $R=W$, and then proceed with the following iterations until the exit condition is met.
\begin{enumerate}
  \item Do SVD of $R$ to get its top-$q$ component of $U, V$, noted as $U^{'}, V^{'}$.
  \item Ternarize $U^{'}, V^{'}$ as a set of row vector and column vector separately, and append them into existing $U, V$.
  \item Calculate $S = \mathcal{\arg\min}_S \|W - U\diag(S)V\|_{F}$.
  \item Calculate $R = W - U\diag(S)V$.
\end{enumerate}

The ternarization policy in step 2 can be summarized as follows: \emph{Given a unit vector $x$, the goal is to find the most sparse ternary vector $x^{*}_{T}$ that satisfies the constraint that the angle between $x$ and ternary vector $x_T$ is less than or equal to a certain threshold $\theta$}. It is important to note that $x^{*}_{T}$ has a specific structure (proved in section \ref{sec:tran_struct}):
\begin{equation}\label{eq:tran_struct}
  \begin{aligned}
    x^{*}_{T}[i] &= 0 \text{ or }\text{sign}(x[i]), \forall i \\
    \arg\text{nonzero}(x^{*}_{T}) &= \arg\text{top\_k}(\text{abs}(x), q), \exists q
  \end{aligned}
\end{equation}
Therefore, instead of traversing all possible $3^{\#(x)}$ Ternary vectors, we only need to traverse $\#(x)$ ternary vectors to find the optimal ($\#(x)$ denote the dimensions of $x$). The complete ternarization algorithm can be viewed in algorithm \ref{alg:binary_policy}.
\begin{algorithm}[htb]
  \DontPrintSemicolon
  \KwIn{$x$, $\theta$\tcp{$x$ is vector, $\theta$ is angle threshold.}}
  $o_x \longleftarrow$ np.sort(np.abs($x$))[::-1] \\
  $s_x \longleftarrow$ np.cumsum($o_x$) \\
  $n_x \longleftarrow$ np.sqrt(np.arange(1, $x$.size+1)) \\
  $c_x \longleftarrow s_x/n_x$ \\
  $i_x \longleftarrow $ np.argmax($c_x \geq \cos(\theta)$) \\
  \If{$c_x[i_x] < \cos(\theta)$}{
      raise error
      \tcp{$T_x$ does not exist.}
  }     
  $t_x \longleftarrow $ np.where($x \geq 0$, 1, -1) \\
  $t_x \longleftarrow $ np.where(np.abs($x$) < $o_x$[$i_x$], 0, $t_x$) \\
  \KwOut{$t_x$ as the ternarizaion of $x$.}
  \caption{Ternarization algorithm $T(\cdot)$(in numpy style), return the most sparsity ternary vector in the set in which angle to $x$ is less or equal to $\theta$.\label{alg:binary_policy}}
\end{algorithm}

To establish the existence of $x^{*}_{T}$, we rely on the following theory (for a more detailed explanation, please refer to section \ref{sec:theta_exist}):
\begin{theorem}\label{thm:theta_exist}
  Considering an $N$-dimensional unit vector $a \in \mathbb{S}^{N-1}$, there must exist a ternary vector $t \in \{\pm 1, 0\}^{N}$ to ensure that their angle $\theta^{'}$ satisfies:
  \[
      \cos(\theta^{'}) = \frac{\langle a, t\rangle}{\sqrt{\|t\|_{0}}} \geq \underbrace{\frac{1}{\sqrt{\sum_{k=1}^{N} (\sqrt{k} - \sqrt{k-1})^2}}}_{\gamma_N}
  \]
\end{theorem}
\begin{remark}
  It can be inferred that the threshold $\gamma_N$ decreases at a rate of $\frac{1}{\sqrt{\log(N)}}$. Although it still approaches zero while $N$ approaches infinity, it decreases very slowly and can reach a minimum value of $\gamma_N \geq \cos(\frac{\pi}{4})$ when $N \leq 55$. Moreover, in practice, we need not worry about the existence of $x^{*}_{T}$ even if we set the threshold $\theta >= \cos(\frac{\pi}{4})$ when $N > 55$ (see section \ref{sec:theta_choice}). This may be related to the specific parameter distribution and requires further exploration.
\end{remark}

For step 3, it is an obvious least squares problem and the optimal solution for $S$, denoted as $S^*$, can be obtained by the following equation (for more details, see section \ref{sec:lsq_for_s}): 
\begin{equation}\label{eq:lsp_for_s}
  S^* = [(U^\top U) \odot (V V^\top)]^{\dagger}\diag(U^\top W V^\top)
\end{equation}
where $\odot$ denotes the Hadamard product and $(\cdot)^{\dagger}$ denotes the Moore-Penrose pseudo-inverse. The entire direct transition algorithm to TSVD form is shown in algorithm \ref{alg:binary_svd}.

\begin{algorithm}[htb]
  \DontPrintSemicolon
  \KwIn{$W$, $\theta$, $q=1$, $R_0=W$, $U_0=\emptyset$, $V_0=\emptyset$}
  $i \longleftarrow 0$ \\
  \While{not exit condition meet }{
      $u, \_, v \longleftarrow$ np.linalg.svd($R_i$, full\_matrices=False)\\
      $T_u \longleftarrow$ vmap($T(\cdot, \cdot)$, in\_axes=(1, None), out\_axes=1)(u[:, 0:$q$], $\theta$) \\
      $T_v \longleftarrow$ vmap($T(\cdot, \cdot)$, in\_axes=(0, None), out\_axes=0)(v[0:$q$, :], $\theta$) \\
      \tcp{Ternarize the top-$q$ column vector of $u$, and top-$q$ row column vector of $v$. See algorithm \ref{alg:binary_policy}.}
      $U_{i+1} \longleftarrow $ np.concatenate([$U_i$, $T_u$], axis=1) \\
      $V_{i+1} \longleftarrow $ np.concatenate([$V_i$, $T_v$], axis=0) \\
      $S_{i+1} \longleftarrow [(U_{i+1}^\top U_{i+1}) \odot (V_{i+1} V_{i+1}^\top)]^{\dagger}\diag(U_{i+1}^\top W V_{i+1}^\top)$ \\
      $R_{i+1} \longleftarrow W - U_{i+1} \diag(s_{i+1}) V_{i+1}$\\
      $i \longleftarrow i+1$ \\
  }
  \KwOut{$U_i, S_i, V_i$, where $U_i \diag(S_i) V_i\ \simeq W $}
  \caption{Direct transition algorithm to TSVD form\label{alg:binary_svd}}
\end{algorithm}

For convergence in theory, we propose theorem \ref{thm:convergence} (proved in section \ref{sec:convergence}):
\begin{theorem}\label{thm:convergence}
  Considering a single iteration in algorithm \ref{alg:binary_svd}, and assuming that $\theta > \frac{\pi}{4}$ and $q=1$, we can define $\bar{S}$ as the argument that minimizes $|R_i - T_u \diag(S) T_v|_F$. In this case, the singular vector of $R_i$ (sorted in descending order) is denoted by $\sigma_i$, and follows that:
  \[
  \begin{aligned}
      \|R_i - T_u \diag(\bar{S}) T_v\|_F^2 & \leq \sin^2(2\theta)(\sigma_i[0])^2 + \|\sigma_i[1:]\|^2_2 \\
                                           & \leq \left( 1 - \frac{\cos^2(2\theta)}{\min(M, N)}\right) \|R_i\|^2_F    
  \end{aligned}
  \]
\end{theorem}

\begin{corollary}
  In fact, compared to the default algorithm \ref{alg:binary_svd}, theorem \ref{thm:convergence} proves convergence on a weaker update policy of $s_i$ by fixing the exist dimensions and just optimizing on the new dimension on the tail. More specifically, when $\theta>\frac{\pi}{4}, q=1$, such weaker update policy is:
  \[
  \begin{aligned}
      \bar{S} &\longleftarrow \arg\min_s \|R_i - T_u \diag(S) T_v\|_F \\
      S_{i+1} &\longleftarrow \textnormal{np.concatenate}([S_i, \bar{S}], \textnormal{axis}=0)
  \end{aligned}
  \]
  This is because under this policy, it must have
  \[
      R_{i+1} = R_i - T_u \diag(\bar{S}) T_v
  \]
  Hence, we can infer that algorithm \ref{alg:binary_svd} must converge at least linearly at a rate of $\sqrt{1 - \frac{\cos^2(2\theta)}{\min(M, N)}}$ when $\theta>\frac{\pi}{4}, q=1$.
\end{corollary}

\begin{remark}
  We still employ a stronger update policy in algorithm \ref{alg:binary_svd} for $S_i$ to prevent the duplication vectors in $U_i$ and $V_i$. Additionally, we have relaxed the constraint for $q > 1$ in consideration acceleration.
\end{remark}

\subsection{Optimal $\theta$ in Practice} \label{sec:theta_choice}

\begin{figure*}[tb]
  \centering
  \begin{subfigure}{.49\textwidth}
    \centering
    \includegraphics[width=\linewidth]{"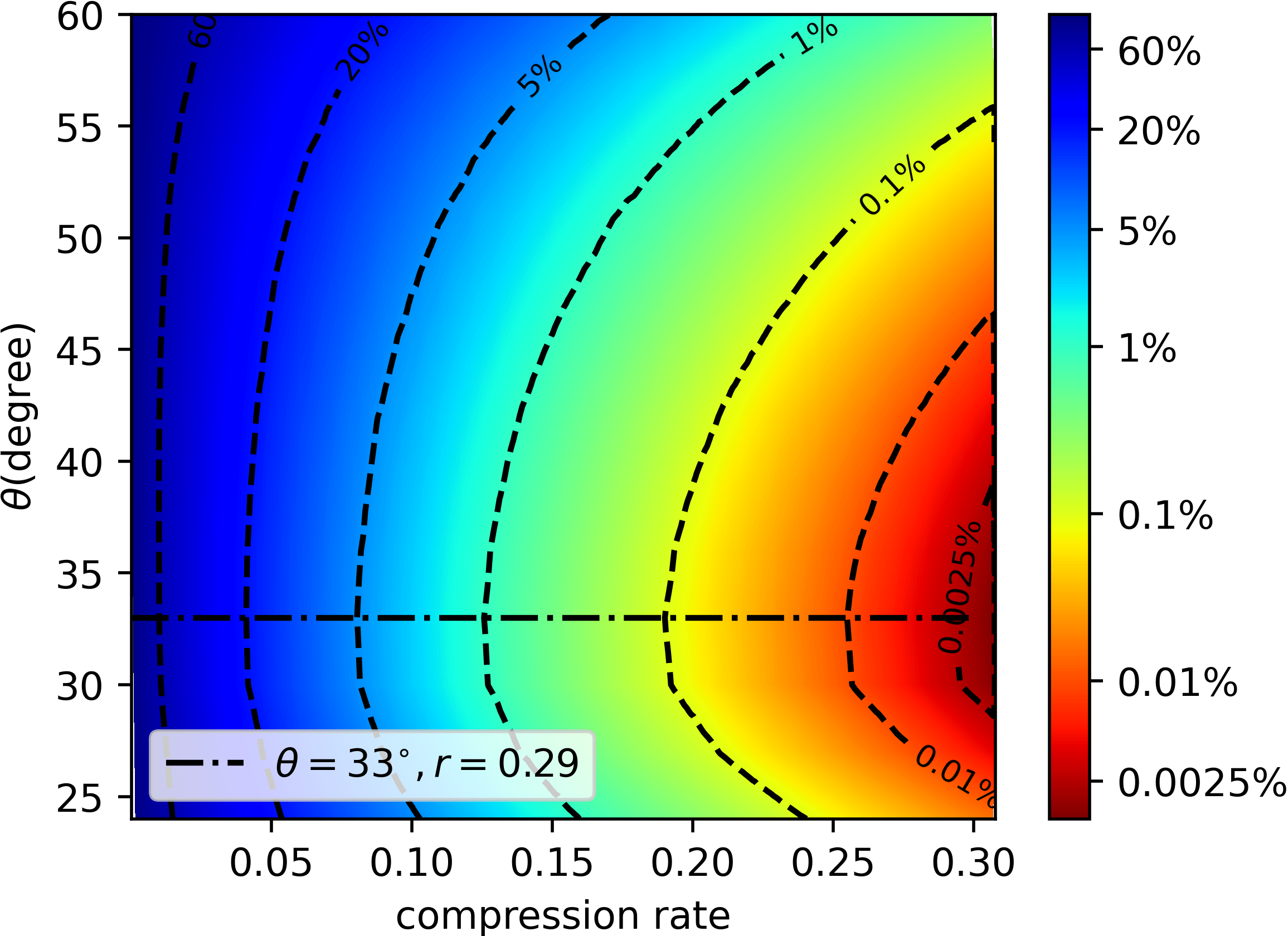"}
    \caption{}\label{fig:tradeoff}
  \end{subfigure}
  \begin{subfigure}{.5\textwidth}
    \centering
    \includegraphics[width=\linewidth]{"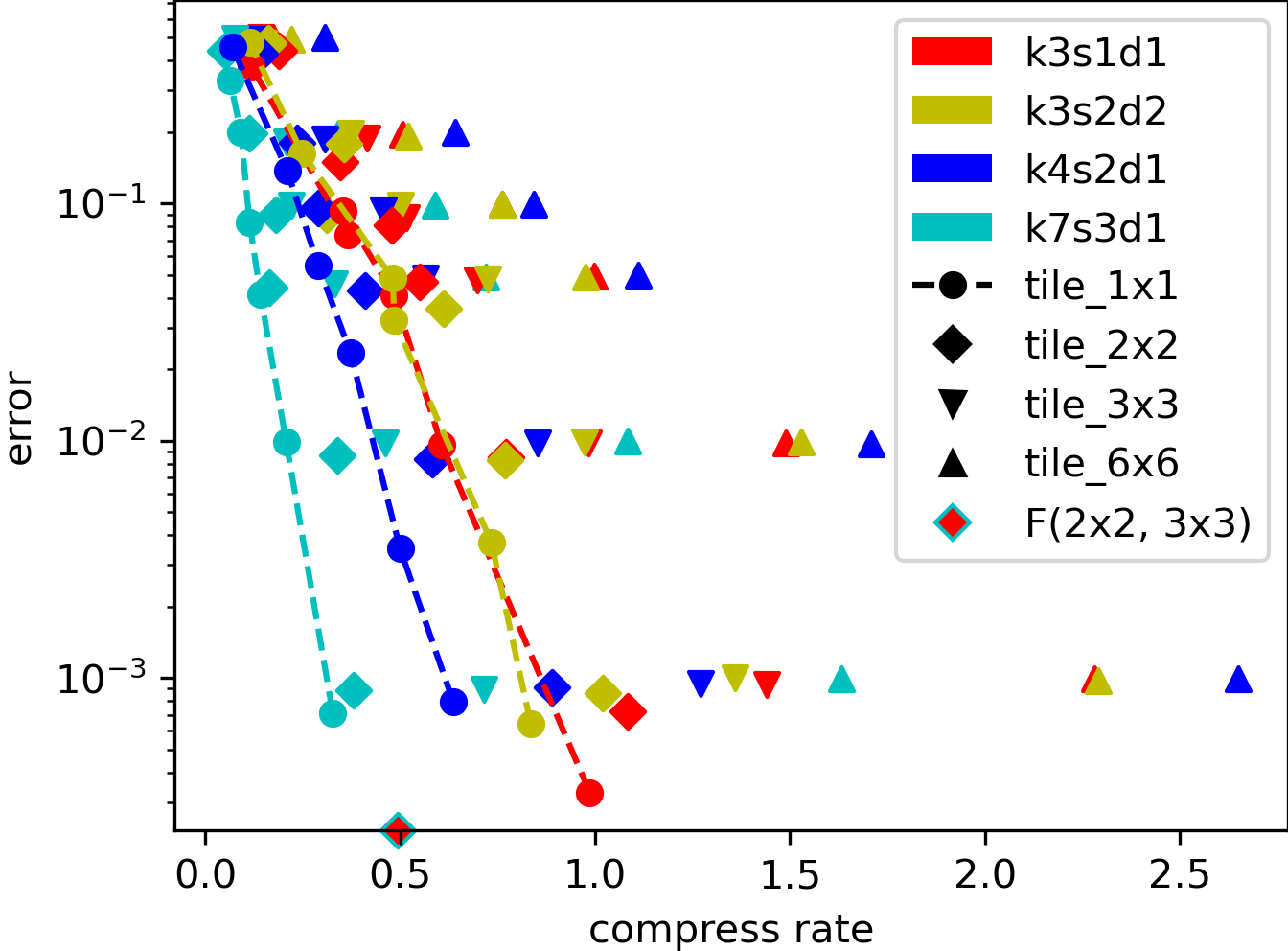"}
    \caption{}\label{fig:conv_select}
  \end{subfigure}
  \caption{(a): The optimal threshold $\theta$ selection on a randomly generated float32 matrix with shape $[512, 256]$ and Laplace distribution. The color in the graph represents the relative error of operator norm. We observed that to achieve the optimal compression rate, the optimal $\theta$ value is approximately $0.576$ (equivalent to $33$ degrees). At this threshold, the sparsity rate $r$ is about 0.29.(b): Tradeoff between compression rate, and relative operator norm error for various single-channel float32 convolutions with different tile unfoldings. Each convolution is represented by a different color, and each tile configuration is represented by a different marker shape. We also highlighted the Winograd $F(2\times 2, 3\times 3)$ in our figure. More experiment detail see section \ref{sec:conv_select}.}
\end{figure*}

To determine the optimal value of $\theta$ in practice, we analyzed the relationship among the relative operator norm error, compress rate, and the $\theta$ threshold. Our testing was conducted on a $[512, 256]$ float32 random matrix with Laplace distribution, and the results are presented in figure \ref{fig:tradeoff}. We found that the optimal value of $\theta$ is achieved when $\theta=0.576(33^{\circ})$ for any tolerance error configuration. Under such $\theta$ the sparsity rate $r$ is about 0.29. These configuration and sparsity rate are stable for various tasks and bit-widths. In this paper, unless otherwise specified, all experiments will be conducted using this configuration and will not additionally report sparsity rate by default.

\subsection{TSVD in Convolution Layer}

For convolution, the most intuitive way to use TSVD is to unfold the kernel into a small tile as a cyclic matrix and process it as a normal linear map. For instance, we can use the Winograd symbol and consider a one-dimensional convolution of $F(2, 3)$, which can be unfolded as follows:
\[
    \left(\begin{aligned} 
        y_1 \\ 
        y_2 \\
     \end{aligned}\right)=\underbrace{\left(\begin{aligned} 
        w_1\  & \quad w_2 & w_3 &\  \quad 0 \\ 
        \quad 0  \  & \quad w_1 & w_2 &\  \quad w_3
    \end{aligned}\right)}_{\mathcal{W}}
    \left(\begin{aligned} 
       x_1 \\ 
       x_2 \\
       x_3 \\
       x_4 \\   
    \end{aligned}\right)
\]
Then, we can perform TSVD on $\mathcal{W}$. In figure \ref{fig:conv_select}, after testing various configurations of tile, stride, dilation, and kernel size, we found that when taking into account the sparsity of $\mathcal{W}$ itself, it has:
\begin{itemize}
  \item  For each convolution configuration, all points lie on the top-right side of the curve for tile\_1x1 in figure \ref{fig:conv_select}, indicating that tile\_1x1 is the optimal tile configuration for convolution TSVD. 
  \item For tile\_1x1, larger kernel size yields a better compression rate, suggesting that larger kernels are easier to compress. 
  \item Convolution TSVD can achieve a $4\%$ error when the compression rate is equivalent to that of $F(2\times 2, 3\times 3)$
\end{itemize}

Therefore, we can simplifies the algorithm as the previous convolution low rank decomposition method does. According to the work of \cite{zhang2015accelerating,jaderberg2014speeding}, we both consider channel-wise and spatial-wise decompositions, i.e. we reshape the kernel into one of the four shapes in eq \eqref{eq:four_shape} and select the optimal one of the compress rate as the result of TSVD in the convolution layer.
\begin{equation}\label{eq:four_shape}
  \begin{aligned}
      \lbrack C_{out}, C_{in} \times K_1 \times K_2 \rbrack \\
      \lbrack C_{out} \times K_1 \times K_2, C_{in} \rbrack \\
      \lbrack C_{out} \times K_1, C_{in} \times K_2 \rbrack \\
      \lbrack C_{out} \times K_2, C_{in} \times K_1 \rbrack
  \end{aligned}
\end{equation}

After decomposition, $U(\cdot)$ and $V(\cdot)$ can be computed in convolution form, and $\diag(s)(\cdot)$ can be calculated as a channel-wise multiplication operator. The kernel shape corresponding to the four shape matrices in eq \eqref{eq:four_shape} is:
\begin{equation*}
  \begin{aligned}
      \text{kernel}_{U} &= \lbrack 1\, 1 \rbrack, &\ \text{kernel}_{V} &= \lbrack K_1, K_2 \rbrack \\
      \text{kernel}_{U} &= \lbrack K_1, K_2 \rbrack, &\ \text{kernel}_{V} &= \lbrack 1, 1 \rbrack \\
      \text{kernel}_{U} &= \lbrack K_1, 1 \rbrack, &\ \text{kernel}_{V} &= \lbrack 1, K_2 \rbrack \\
      \text{kernel}_{U} &= \lbrack 1, K_2 \rbrack, &\ \text{kernel}_{V} &= \lbrack K_1, 1 \rbrack
  \end{aligned}
\end{equation*}

\subsubsection{Connection with Winograd}

It can be inferred that the most commonly used Winograd convolution form, $F(2 \times 2, 3	\times 3)$, is actually \emph{ one of the special lossless cases of TSVD on a tile of size $2 \times 2$}. As shown in figure \ref{fig:conv_select}, our algorithm with a tile size of $1\times1$ can achieve a better compression rate than Winograd $F(2	\times 2, 3	\times 3)$ when tolerance error is great or equal to $4\%$. Another advantage of decomposing on a tile size of $1 \times 1$ is that our algorithm can adapt to any configurations of stride, padding, and dilation. Even on depth-wise convolution with a large kernel, our algorithm can obtain a pretty good compression rate, giving us greater flexibility compared to fixed decomposition methods like Winograd.

\subsection{Training TSVD in QAT style}\label{sec:QAT}

\begin{algorithm}[htb]
  \DontPrintSemicolon
  \KwIn{$W^{t+1}$, $U^{t}$, $V^{t}$, $\eta$, $\theta$}\tcp{$\eta$ is the threshold of main-tail split.}.
  $S \longleftarrow [(U^{t\top} U^{t}) \odot (V^{t} V^{t\top})]^{\dagger}\diag(U^{t\top} W^{t+1} V^{t\top})$ \\
  $R \longleftarrow W - U^{t} \diag(S) V^{t}$\\
  $U', S', V' \longleftarrow \mathop{\text{TSVD}}(R, \theta, rank=1)$\\
  \tcp{get the maximum TSVD singular of the residual.}
  $m \longleftarrow S\sqrt{\diag(U^{t\top} U^{t}) \diag(V^{t} V^{t\top})} > \eta S'\|U'\|\|V'\| $\\
  $U, V \longleftarrow U^{t}[\dots, m], V^{t}[m, \dots]$ \\
  $S \longleftarrow [(U^{\top} U) \odot (V V^{\top})]^{\dagger}\diag(U^{\top} W^{t+1} V^{\top})$ \\
  $R \longleftarrow W - U \diag(S) V$\\
  \tcp{recompute the residual after determine the main part.}
  $U^{t+1}, S^{t+1}, V^{t+1} \longleftarrow \mathop{\text{TSVD}}(W^{t+1}, \theta, R_0=R, U_0=U, V_0=V) $ \tcp{algorithm \ref{alg:binary_svd}}

  \KwOut{$U^{t+1}, S^{t+1}, V^{t+1}$}
  \caption{Recompute policy in TSVD QAT training.\label{alg:QAT}}
\end{algorithm}

For very low bit training, including BNN, the key essential of QAT training or finetuning is how to back-propagate through the quantized parts. To achieve this, we still use the Straight-Through Estimator (STE) for TSVD QAT training but in a novel way. \emph{Instead of considering how to back-propagate to $U, S$ and $V$, we view the entire TSVD result of $W$ as a quantization form of $W$ and apply STE to it.} Specifically, we perform this process during forwarding:
\[
    \left\{\begin{aligned}
      U, S, V &= \text{TSVD}(W) \\
      \bar{W} & =  U\diag(S)V \\
      y &\simeq \bar{W}x
    \end{aligned}\right.
\]
and do such STE in back-propagation:
\[
    \frac{\partial \mathcal{L}}{\partial W} \simeq \frac{\partial \mathcal{L}}{\partial \bar{W}}
\]

Hence, we need to redo the TSVD of $W$ every time the optimizer updates the parameters. This task is computationally intensive, so we only hold the main row and column vectors of $V$ and $U$ corresponding to the top $S$ values, and recompute their tail part. This approach is outlined in algorithm \ref{alg:QAT}. For the convolution layer, the matrix type in eq \ref{eq:four_shape} is only reselected when the main row and column vectors of $V$ and $U$ are empty.

\section{Experiment}
We test TSVD on various scales of neural networks and tasks, consistently setting $\theta = 0.576$ as demonstrated in section \ref{sec:theta_choice}, setting a self-adaptive $q$ to guarantee at least 20 iterations in algorithm \ref{alg:binary_svd}. The only hyperparameter in our algorithm that requires tradeoff is the tolerance error. We follow the method presented in table \ref{tab:compute_cost} for the translation of acceleration rate of all comparative items. Due to computational limitations, we only present the finetune results of the image task. All experiments are conducted on a single GPU node equipped with $8\times$ Nvidia Tesla V100 32G GPUs, and the results are presented as the mean of three runs.
\subsection{Experiment on ImageNet 1K}\label{sec:imagenet_1k}

We test ConvNeXt-T \cite{liu2022convnet}, Swin-T \cite{liu2021swin} and ResNet-50 \cite{he2016deep} on ImageNet 1K \cite{imagenet15russakovsky} dataset and showed in table \ref{tab:exp_imagenet}. One should be highlight is that different to the previous compression method like \cite{chen2020addernet} or \cite{li2021towards}, our method has enough accuracy hence we transform \emph{all linear and convolution layers in those networks, including the convolution layers in stem, fully connect layer in head, and all depth-wise convolution}. Note these three network architecture are quite different: ResNet50 is the classic baseline with $3\times 3$ convolution, Swin-T is the compact visual transform architecture, and ConvNeXt-T is composed by $1\times 1$ convolution with large channel and depth-wise convolution with large kernel. TSVD can achieve high acceleration rate with almost lossless accuracy drop on all three models. The final sparsity rate and rank $K$ for each layer can be viewed in section \ref{sec:imagenet_1k_detail}

\begin{table*}[htb]\tiny 
	\centering\caption{ TSVD tests on Imagenet 1k using various networks. The acceleration rate estimated by $d$ bit-width is expressed as acc($d$), although every network was actually run under float32. $\times(B)$ denote the multiplication billion counts, $+(B)$ denote the addition billion counts, $\times +(B)$ denote the  multiply-accumulate billion counts. F denote finetuning.}\label{tab:exp_imagenet}
	\resizebox{\textwidth}{10\baselineskip}{
    \begin{tabular}{|c c|c c c c|c c|c|}
		\hline
		Model & Method & $\times$(B) & $+$(B) & acc($d$=32) & acc($d$=8) & params (M) & tern params (M) & top-1(\%)\\\hline
    ConvNeXt-T & original & 4.47 & 4.46  & $\times$1 & $\times$1 & 28.6 & 0 & 82.07 \\
    & \textbf{$1\%$ tol TSVD} & 0.074 & 12.6 & $\times$9.34 & $\times$2.40 & 0.23 & 279.3 & \textbf{82.05}\\
    & \textbf{$7\%$ tol TSVD (F)} & 0.046 & 6.12  & \textbf{$\times$18.47} & \textbf{$\times$4.89} & 0.15 & 129.78 & 82.04\\
    \hline
    Swin-T & original & 4.50 & 4.50  & $\times$1 & $\times$1 & 28.3 & 0 & 80.40 \\
    & \textbf{$1\%$ tol TSVD} & 0.21 & 12.3 & $\times$7.50 & $\times$2.32 & 0.29 & 261.8 & \textbf{80.37}\\
    & \textbf{$7\%$ tol TSVD (F)} & 0.18 & 5.84 & \textbf{$\times$12.41} & \textbf{$\times$4.55} & 0.19 & 109.03 & 80.34\\
    \hline
    ResNet-50 & original & 4.10 & 4.10  & $\times$1 & $\times$1 & 25.6 & 0 & 75.85 \\
    & \textbf{$1\%$ tol TSVD} & 0.060 & 10.83 & $\times$10.06 & $\times$2.56 & 0.21 & 242.5 & \textbf{75.81}\\
    & \textbf{$7\%$ tol TSVD (F)} & 0.035 & 4.99 & \textbf{$\times$21.04} & \textbf{$\times$5.51} & 0.16 & 105.23 & 75.79\\
    & AdderNet\cite{chen2020addernet} & 0.131 & 8.08 & $\times$10.58 & $\times$3.24 & 25.6 & 0 & 74.94\\\cline{3-8}
    & & $\times +$ (B) &  & acc & & params (M) & & \\\cline{3-8}
    & CC($C=0.5$)\cite{li2021towards} & 1.93 &  & $\times$2.12 & & 13.2 & & 75.59\\
    & LRPET($P=0.62$)\cite{guo2022compact} & 1.90 &  & $\times$2.15 & & 12.89 & & \textbf{75.91}\\
    \hline
	\end{tabular}}
\end{table*}

\subsection{Experiment on BERT}

We run experiment on our TSVD method using the BERT base model \cite{devlin2018bert} and the GLUE dataset \cite{wang2018glue}. As in section \ref{sec:imagenet_1k}, we applied TSVD to the last fully connected head layer, but not to the embedding layer, as it does not contain any FLOPS. The sequence length was fixed at 128, and for each downstream task in GLUE, we first fine-tuned the model as usual, then performed a direct TSVD transition at the end. The computation cost after TSVD for various downstream tasks was similar, so we only report the mean of multiplication, addition, and parameter count of these downstream tasks. Compared to the original BERT base, TSVD achieved an acceleration rate of $\times 8.75$ with almost no loss in accuracy, and $\times 13.45$ with only a slight loss in accuracy, which exceeds the current state-of-the-art method, such as prune OFA \cite{zafrir2021prune}.

\begin{table*}[htb]\tiny 
	\centering\caption{BERT experiment on GLUE dataset. The acceleration rate estimated by $d$ bit-width is expressed as acc($d$), although every network was actually run under float32. P / TP (M) represents the standard parameters and ternary parameters in millions. $\times / + (B)$ denote the multiplication and addition billion counts, $\times +(B)$ denote the  multiply-accumulate billion counts.}\label{tab:exp_glue}
	\resizebox{\textwidth}{5\baselineskip}{
    \begin{tabular}{|c c|c c|c|c c c c c c c c|}
		\hline
		\multirow{2}{*}{Model} & \multirow{2}{*}{Method} & \multirow{2}{*}{$\times$/+(B)} &  \multirow{2}{*}{acc($d$=32 / 4)} & \multirow{2}{*}{P / TP (M)} & CoLA & SST-2 & MRPC & STS-B & QQP & MNLI & QNLI & RTE \\
     & & & & & Matthews Corr & Accuracy & F1 / Accuracy & Pearson / Spearman & F1 / Accuracy & M / MM & Accuracy & Accuracy \\
    \hline
    BERT & original & 11.19 / 11.18  & $\times$1 / 1 & 109 / 0 &  59.33 & 92.78 & 89.19 / 84.55 & 87.52 / 87.23 & 87.50 / 90.81 & 83.79 / 84.27 & 90.61 &64.26 \\
      base & \textbf{$1\%$ tol TSVD} & 0.34 / 29.40 &  $\times$8.75 / 1.47 & 23 / 825 & \textbf{60.81} & \textbf{92.43} & \textbf{89.03 / 83.57} & \textbf{88.47 / 88.28} & \textbf{87.42 / 90.68} & \textbf{83.50 / 84.36} & \textbf{90.57} & \textbf{65.70} \\
    & \textbf{$5\%$ tol TSVD} &  0.33 / 15.88 & \textbf{$\times$13.45 / 2.65} & 23 / 440  & 60.65 & 91.05 & 89.78 / 85.04 & 87.57 / 87.40 & 86.71 / 89.51 & 83.11 / 82.75 & 89.36 &61.01 \\\cline{3-5}
    & & $\times +$ (B) & acc & params(M) & & & & & & & & \\\cline{3-5}
    & 85\% prune OFA \cite{zafrir2021prune} & 1.94 & $\times$ 5.76 & 109 & 41.51 & 90.48 & 87.25 / 82.60 & 82.86 / 83.13 & 84.44 / 88.53 & 78.89 / 79.53 & 88.01 & 54.51 \\
    & 90\% prune OFA \cite{zafrir2021prune} & 1.40 & $\times$ 7.99 & 109 & 35.60 & 88.76 & 83.90 / 75.74 & 81.53 / 82.08 & 83.73 / 87.84 & 77.61 / 78.36 & 86.91 & 56.32 \\
    \hline
	\end{tabular}}
\end{table*}

\subsection{Experiment on OPT}

To evaluate our method on large language models, we conducted experiments on all linear layers of OPT-6.7B \cite{zhang2022opt} with a full sequence length of 2048, including the last lm\_head layer. As shown in Table \ref{tab:exp_opt}, our method achieved an acceleration rate of approximately $\times 4.64$ and produced the most lossless results on wikitext2 \cite{merity2016pointer}, ptb \cite{dinarelli2019seq2biseq}, and c4 \cite{2019t5}. Our acceleration rate is better than the current state-of-the-art methods such as GPTQ \cite{frantar2022gptq} and sparseGPT \cite{frantar2023sparsegpt}, while our method does not have an advantage in parameter storage. The acceleration rate calculation method follows the approach outlined in table \ref{tab:compute_cost}.

\begin{table*}[htb]\tiny 
	\centering\caption{Experiment on the language generation model OPT-6.7B with a sequence length of 2048, evaluating the acceleration rate when $d=16$.The notation $\times$(T) represents trillion multiplications counts, while +(T) represents trillion additions counts. $\times +$(T) denotes trillion multiply-accumulate operations counts with a specific weight and activation bit width. The indicator of datasets wikitext2, ptb and c4 is Perplexity (PPL).}\label{tab:exp_opt}
	\resizebox{\textwidth}{6\baselineskip}{
    \begin{tabular}{|c c|c c c|c c|c c c|}
		\hline
		Model & Method & $\times$(T) & +(T) &  acc($d$=16) & params (B) & tern params (B) & wikitext2 & ptb & c4 \\\hline
    OPT & original & 14.72 & 14.72  & $\times$1 & 6.86 & 0 &  10.86 &  13.08 & 11.74 \\
    6.7b & \textbf{$1\%$ tol TSVD} & 1.11 & 31.98 &  $\times$4.64 & 0.22 & 55.03 & \textbf{11.10} & \textbf{13.73} & \textbf{12.16} \\ 
    & \textbf{$1.5\%$ tol TSVD} & 1.11 & 27.66 &  $\times$5.11 & 0.22 & 47.37 & 12.12 & 15.62 & 13.34\\ 
      & \textbf{$2\%$ tol TSVD} & 1.11 & 24.64 &  \textbf{$\times$5.49} & 0.22 & 42.00 & 19.08 & 26.06 &  25.75 \\\cline{3-7}
    & & W16A16, $\times +$ (T) & W$d'$A16, $\times +$(T) &  acc & params (B) & $d'$-bit params (B) & & & \\\cline{3-7}
    & 4-bit GPTQ & 1.53 & 13.19 &  $\times$3.75 & 0.43 & 6.41 & 11.39 & 13.77 &  \textbf{12.14} \\
    & 3-bit GPTQ & 1.53 & 13.19 &  $\times$4.66 & 0.43 & 6.41 & 14.98 & 18.67 &  15.54 \\
    & 50\% sparseGPT  & 8.13 & 0 &  $\times$1.81 & 6.86 & 0 & 11.59 & 17.38 &  13.72 \\
    & 50\% sparseGPT + 4 bit & 1.53 & 6.59 &  $\times$5.16 & 0.43 & 6.41 & 12.23 & 18.16 &  14.22 \\
    \hline
	\end{tabular}}
\end{table*}

\section{Conclusion and Discussion}

In this paper, we introduce Ternary SVD as an improved parameterized form of linear mapping. By combining low-bit quantization and SVD, Ternary SVD effectively vanishes the number of multiple instructions in fully connected and convolution layers. We provide direct and training transition algorithms for Ternary SVD like Post Training Quantization and Quantization Aware Training respectively. Additionally, we analyze the convergence of the direct TSVD transition algorithms in theory. Our experiments demonstrate that Ternary SVD achieves state-of-the-art network compression performance across various networks and tasks, including current baseline models like ConvNext, Swim, BERT, and large language model like OPT.

However, there are still some bottlenecks that prevent the out-of-the-box usage of TSVD. The main issue is that although the compute cost is theoretically $\mathcal{O}(d)$ for $d$-bit addition instructions and $\mathcal{O}(d^2)$ for $d$-bit multiplication instructions separately, as far as we know, the main vectorized computation platforms like CUDA or MKL do not optimize for it and still assign the same instruction cycle. Therefore, in practice, it is currently unavailable to achieve the wall-time advantage of TSVD as guaranteed by theory on the mainstream hardware platforms. This is also the reason why we only report the counts of addition and multiplication instructions and do not show the wall-time latency in our experiments.

Another issue is the sparsity acceleration problem. While sparsity is an attractive feature in TSVD, it is not friendly in vectorized computation and has been a study for many years in research. However, recent advancements in vectorized computation have proposed structured sparsity instructions, e.g. in \cite{mishra2021accelerating} the 2:4 sparse tensor core in Nvidia A100 GPU. One of our future works is to further constrain TSVD to satisfy the structure of these instructions.

For large language models, although our method achieves remarkable acceleration performance compared to other low-bit compression algorithms, it does not provide an advantage in terms of parameter storage. However, we are relatively optimistic about this issue because the ternary matrix $U$ and $V$ are quite sparse, and the current 2-bit presentation is overly redundant. Presenting in a more efficient way is another future work in our plan.

Additionally, we would like to discuss the connection between BNN and its potential evolution in the end of this paper.
\subsection{Connection with BNN and What Next}
In a sense, TSVD can also be seen as the weight component of the BNN PTQ transition. This is due to the fact that we can decompose $U$ and $V$ into their positive and negative parts:
\begin{equation*}
    \left\{\begin{aligned}
        U &= U_{+} - U_{-} \\
        V &= V_{+} - V_{-} \\
    \end{aligned}\right.
\end{equation*}
where $U_{+}, U_{-}, V_{+}, V_{-}$ are binary matrices in $\{0, 1\}$. One important thing to highlight is that \emph{such decomposition is not only mathematically equivalent, but also equivalent in terms of FLOPS counts and parameter storage.} The equivalence in parameter storage is due to that we only need 2 bits to store each element of a ternary matrix, and 1 bit of a binary matrix. The equivalence in FLOPS counts is due to that if there is a $\pm 1$ element in $U$, there should be only one 1 in $U_{+}$ or $U_{-}$ at the corresponding index.

Therefore, a natural and appealing idea is to further develop the BNN PTQ method by exploring how to binarize the feature map in the middle layer of networks. However, whether this can be achieved or not, such idea is not a proper question. This is because the computational cost of $d$-bit addition instruction is $\mathcal{O}(d)$, and \emph{the computation cost in network layers dominated by addition is always proportional to the total bits of the input feature map, regardless of the bit width in computation}. Therefore, transitioning to a binary feature map is unhelpful if the total bits of the input feature map do not decrease.

In other words, rather than solely focusing on binarization, the focus should shift towards \emph{improving the coding efficiency of the feature map and reducing the overall bit count of feature map after TSVD transition}. This might be a more challenging task compared to pure binarization.

\bibliographystyle{aaai24}
{ \small
  \bibliography{./bib/base}
}

\newpage
\vfill\null
\newpage
\begin{center}
\textbf{\large Supplemental Materials}
\end{center}

\setcounter{equation}{0}
\setcounter{figure}{0}
\setcounter{table}{0}
\setcounter{page}{1}
\setcounter{section}{0}
\makeatletter
\renewcommand{\theequation}{S\arabic{equation}}
\renewcommand{\thetable}{S\arabic{table}}
\renewcommand{\thefigure}{S\arabic{figure}}
\renewcommand{\bibnumfmt}[1]{[S#1]}
\renewcommand{\citenumfont}[1]{S#1}
\renewcommand{\thesection}{S\arabic{section}}

\section{Proof}
\subsection{Proof of Eq \eqref{eq:tsvd_critical_rank}}\label{sec:tsvd_critical_rank_detail}
\begin{table}[htb]
	\centering\caption{Computing Cost in TSVD}\label{tab:compute_cost}
	\resizebox{\linewidth}{3\baselineskip}{\begin{tabular}{|c|c|c|c|}
		\hline
		operator & $\times$ & $+$ & compute cost(equivalent $+$) \\\hline 
        $W(\cdot)$ & $MN$ & $MN$ & $MN(d-1)$ \\\hline 
        $U(\cdot)$ & - & $rMK$ & $rMK$ \\\hline
        $\diag(s)(\cdot)$ & $K$ & - & $K(d-2)$ \\\hline
        $V(\cdot)$ & - & $rKN$ & $rKN$ \\\hline
        $U\diag(s)V(\cdot)$ & $K$ & $rK(M+N)$ & $rK(M+N) + (d-2)K$ \\\hline
	\end{tabular}}
\end{table}  
\begin{proof}
  The cost of computing each component in TSVD is shown in table \ref{tab:compute_cost}. Therefore, the critical rank $\bar{K}$ should be such that:
  \[
    r\bar{K}(M+N) + (d-2)\bar{K} = MN(d-1)
  \]
  which is equivalent to eq \eqref{eq:tsvd_critical_rank}.
\end{proof}
\subsection{Proof of Eq \eqref{eq:tran_struct}}\label{sec:tran_struct}

We notice an obvious lemma:
\begin{lemma} \label{lem:tran_struct} 
  For any $x \in \mathbb{R}^N$, and any ternary vector $x_T \in \{ \pm 1, 0 \}^N$, we define $n$ as the count of non-zero elements in $x_T$. We can then construct a ternary vector $x'_T$ such that for any $i \in [0, N)$: 
  \[
    x'_T[i] = \left\{\begin{aligned}
        & \text{sign}(x[i])\text{ if abs}(x[i])\text{ is the top-}n\text{ elements of abs}(x) \\
        & 0, \text{else}
    \end{aligned}\right.
  \]
  It follows that: 
  \[\begin{aligned}
    \langle x'_T, x \rangle &\geq \langle x_T, x \rangle \\
    \| x'_T \| & = \|x_T\|
  \end{aligned}\]
\end{lemma}
Therefore, if the optimal value of $x^*_T$ does not satisfy eq \eqref{eq:tran_struct}, we can utilize lemma \ref{lem:tran_struct} to construct a superior $x^{*'}_T$. However, this contradicts the optimality of $x^*_T$.

\subsection{Proof of Theorem \ref{thm:theta_exist}}\label{sec:theta_exist}

We proof the following lemma:
\begin{lemma}\label{lem:ineq}
  Assuming $a\in \mathbb{R}_{+}^N$, where $a[0] \geq a[1] \geq \dots \geq a[N-1] \geq 0$ and $\|a\| = 1$, it follows that: 
  \[
      \min_a\max_{1\leq k \leq N} \frac{\sum_{i=0}^{k-1}a[i]}{\sqrt{k}} = \gamma_N
  \] 
  where $\gamma_N$ is defined in theorem \ref{thm:theta_exist}.
\end{lemma}
\begin{remark}
  This lemma is intuitionistic, as it suggests that the minimum value is reached when $\forall k, \frac{\sum_{i=0}^{k-1}a[i]}{\sqrt{k}}$ remains constant. However, a rigorous proof is complex and is demonstrated as below. Once lemma \ref{lem:ineq} is proven, we can easily deduce theorem \ref{thm:theta_exist} by incorporating the structure outlined in lemma \ref{lem:tran_struct}.
\end{remark}
\begin{proof}
  We can divide the feasible region of $a$ into $N$ parts by introducing an additional constraint $C(k'), 1 \leq k' \leq N$. This constraint is defined as follows:
  \[
    \arg\max_{1\leq k \leq N} \frac{\sum_{i=0}^{k-1}a[i]}{\sqrt{k}} = k'
  \]
  under such constraint, it follow that:
  \begin{equation}\label{eq:ineq}
    \min_{a, C(k')}\max_{1\leq k \leq N} \frac{\sum_{i=0}^{k-1}a[i]}{\sqrt{k}} = \min_{a, C(k')} \frac{\sum_{i=0}^{k'-1}a[i]}{\sqrt{k'}}
  \end{equation}
  Now, we can express the optimization problem in equation \eqref{eq:ineq} with $C(k')$ in a standard form:
  \begin{equation}\label{eq:ineq_2}
    \left\{\begin{aligned}
       \min & \frac{1}{\sqrt{k'}}b^\top a &\\
      s.t. & -Q a &\leq 0 & \quad\text{$a$ is decrease by index} \\
           &  (I - L)\diag(v)^{-1} Q^{-1\top} a &\leq 0 & \quad \text{ equivalent to $C(k')$}\\
           & \|a\|^2 - 1 & = 0 & \quad  \\
    \end{aligned}\right.
  \end{equation}
  where $I$ is the identity matrix, $Q, L, v, b$ is defined as follow:
  \begin{align*}
    Q = \left(\begin{aligned}
      1 & -1 &  & & \\
        & \quad 1 & -1 & &  \\
        &  & \ddots & \ddots & \\
        &  & & \quad 1 & -1 \\
        &  &  &  & \quad 1
    \end{aligned} \right), 
    L =\left(\begin{aligned}
      \dots 0& \quad 1 & 0 \dots\\
      \dots 0& \quad 1 & 0 \dots\\
           & \quad \vdots & \\
      \dots 0& \quad 1 & 0 \dots\\
    \end{aligned} \right) \\
    v = {\left(1 \quad \sqrt{2} \quad \dots \quad\sqrt{N}\right)}^{\top},
    b = {\left( 1 \dots 1 \quad 0 \dots 0 \right)}^{\top}
  \end{align*} 
  $L$ has only one vector with all elements equal to 1 on its $k'$-th column, $b$ has 1 on its first $k'$ indices. Therefore, 
  \[
    Q^{-1\top} = \left(\begin{aligned}
      1 &  &  & \\
      1 & \quad 1 &  &  \\
      \vdots  & \quad \vdots & \ddots & \\
      1  &  \quad 1 & \dots & \quad 1 \\
    \end{aligned}\right)
  \]
  Now, we will construct a less restrictive problem in comparison to eq \eqref{eq:ineq_2}
  \begin{equation}\label{eq:ineq_3}
    \left\{\begin{aligned}
      \min & \frac{1}{\sqrt{k'}}b^\top a &\\
      s.t. & -Q a &\leq 0 & \quad\text{$a$ is decrease by index} \\
           &  (I - L)\diag(v)^{-1} Q^{-1\top} a &\leq 0 & \quad \text{ equivalent to $C(k')$}\\
           & \|a\|^2 - 1 & \leq 0 &  \quad \text{relax condition 1}\\
           & 1 - v^\top Q a &\leq 0 & \quad \text{relax condition 2}\\
    \end{aligned}\right.
  \end{equation}
  The reason for the relaxed condition 2 in equation \eqref{eq:ineq_3} is that the set $\{a| -Q a \leq 0\}$ is a convex cone based on the column vector of $Q^{-1}$. Additionally, the function $f$ defined as
  \[  
    f(a) = \|a\| - v^\top Q a
  \] 
  is convex, and for each column vector in $Q^{-1}$, denoted as $a'$, it must satisfy $f(a')=0$. Therefore, 
  \[  
    f(a) = \|a\| - v^\top Q a \leq 0, \forall a \in \{a| -Q a \leq 0\}
  \]
  which is equivalent to relax condition 2 in eq \eqref{eq:ineq_3} after normalization of $\|a\|=1$.
  Noticing that the feasible area of \eqref{eq:ineq_3} is convex, we can apply the Karush-Kuhn-Tucker Conditions \cite{avriel2003nonlinear} to obtain the necessary and sufficient conditions for optimal $a^*$ in eq \eqref{eq:ineq_3}. Specifically, these conditions require the existence of $u$ and $w$ in $\mathbb{R}^N$, as well as $\lambda$ and $\beta$ in $\mathbb{R}$, such that:
  \begin{equation}\label{eq:KKT}
    \left\{\begin{aligned}
      \frac{1}{\sqrt{k'}} b - Q^\top u & + \\ Q^{-1}\diag(v)^{-1}(I - L^\top)w + 2 \lambda a^{*} - \beta Q^\top v &= 0 \\
      u, w, \beta, \lambda &\geq 0 \\
      -Q a^{*} &\leq 0 \\
      (I - L)\diag(v)^{-1} Q^{-1\top} a^{*} &\leq 0\\
      \|a^{*}\|^2 - 1 &\leq 0\\
      1 - v^\top Q a^* &\leq 0 \\
      -u^\top Q a^{*} &= 0 \\
      w^\top (I - L)\diag(v)^{-1} Q^{-1\top} a^{*} &= 0\\  
      \lambda (\|a^{*}\|^2 - 1) &= 0 \\
      \beta(1 - v^\top Q a^*) &= 0
    \end{aligned} \right.
  \end{equation}
 We can verified that $\|Q^\top v\| = \frac{1}{\gamma_N}$ and the solution presented below is a KKT point.
  \[
    \left\{\begin{aligned}
     u &= 0 \\
     w &= \gamma_N^2 \diag(v) Q Q^\top v \\
     a^*& = \gamma_N Q^\top v \\
     \lambda &= \frac{\gamma_N}{2} \\
     \beta &= 2 \gamma_N^2 \\
     \text{eq } \eqref{eq:ineq_3} &= \gamma_N
    \end{aligned} \right.
  \]
  Therefore, it is sufficient to be the minimum point of eq \eqref{eq:ineq_3}, while also being a feasible point of eq \eqref{eq:ineq_2}. As a result, it can be concluded that it must be the minimum point of eq \eqref{eq:ineq_2}, which means:
\[
  \min_{a, C(k')} \frac{\sum_{i=0}^{k'-1}a[i]}{\sqrt{k'}} = \gamma_N
\]
Now, if we traverse all possible values of $k'$, we can obtain:
\begin{align*}
  \min_a\max_{1\leq k \leq N} \frac{\sum_{i=0}^{k-1}a[i]}{\sqrt{k}} = \min_{1\leq k' \leq N}\min_{a, C(k')} \frac{\sum_{i=0}^{k'-1}a[i]}{\sqrt{k'}} = \gamma_N
\end{align*}
\end{proof}

\subsection{Proof of Eq \eqref{eq:lsp_for_s}}\label{sec:lsq_for_s}
\begin{proof}
  We can define a linear mapping $H$ from $S$ to $W$:
  \begin{equation}\label{eq:H_define}
    H: S \rightarrow W, W = H(S) =  U \diag(S) V
\end{equation}
Then, it can be easily obtained as follows:
\begin{align*}
    H^{\top}(W) & = \diag(U^\top W V^\top) \\
    H^{\top}H(S) & = [(U^\top U) \odot (V V^\top)](S)
\end{align*}
Hence, the optimal $S^*$ can be expressed as:
\[
    S^* = (H^{\top}H)^{\dagger}H^{\top}(W) = [(U^\top U) \odot (V V^\top)]^{\dagger}\diag(U^\top W V^\top)
\]
\end{proof}

\begin{table*}[tb]
	\centering\caption{Cost translation method for computing the difference compression method on a linear mapping $W(\cdot)$ with a $d$-bit width. The input is a vector, and $W$ is a matrix with shape $[M, N]$. We only consider weight quantization, specifically in the form of $W_{d'}A_d$.}\label{tab:compute_cost}
	\resizebox{\textwidth}{4\baselineskip}{
    \begin{tabular}{|c|c|c|c|c|c|}
		\hline
		method & hyperparameter & $\times +$ counts & equivalent $+$ & compression rate & acceleration rate \\\hline 
    origin & matrix shape $[M, N]$ & $MN$ & $(d-1)MN$ & 1 & $\times 1$ \\
    SVD &  rank $K$  & $MK + KN$ &$(d-1)(MK + KN)$ & $\frac{MK + KN}{MN}$ & $\times \frac{MN}{MK + KN}$ \\
    puring & sparsity $r$ & $rMN$ &$r(d-1)MN$ & $r$ & $\times \frac{1}{r}$ \\
    quantization & quant form W$d'$A$d$ & $ MN$ & $(d'-1)MN$ & $\frac{d'-1}{d-1}$ & $\times \frac{d-1}{d'-1}$ \\\cline{3-3}
    & & $\times$ / $+$ counts & & &\\\cline{3-3}
    TSVD & sparsity $r$, rank $K$ & $K$ / $rK(M+N)$ & $K(d-2) + rK(M+N)$ & $\frac{K(d-2) + rK(M+N)}{MN(d-1)}$ & $\times \frac{MN(d-1)}{K(d-2) + rK(M+N)}$\\
    \hline
	\end{tabular}}
\end{table*}  

\subsection{Proof of Theorem \ref{thm:convergence}}\label{sec:convergence}
\begin{proof}
  We consider the extended SVD decomposition of $R_i$: $R_i = \mathcal{U} \Sigma \mathcal{V}$, where $\mathcal{U}$ is a unitary matrix with shape $[M, M]$, $\mathcal{V}$ is a unitary matrix with shape $[N, N]$, and $\Sigma$ is a diagonal matrix with shape $[M, N]$. We denote the diagonal vector of $\Sigma$ as $\sigma_i$. We first normalize each column vector in $T_u$ to length 1, denoted as $T^{'}_u$, and normalize each row vector of $T_v$ to length 1, denoted as $T^{'}_v$. Note that $\mathcal{U}$ and $\mathcal{V}$ are fully rank, so we can express $T^{'}_u$ and $T^{'}_v$ on the basis of $\mathcal{U}$ and $\mathcal{V}$ separately: 
  \[
      \begin{aligned}
          T'_u &= \mathcal{U}P_u\\
          T'_v &= P_v\mathcal{V}
      \end{aligned}
  \]
  where $P_u = \mathcal{U}^\top T'_u, P_v =T'_v \mathcal{V}^\top$ are two matrices with shape $[M, 1]$ and $[1, N]$, respectively. From algorithm \ref{alg:binary_policy}, we can obtain:
  \[
    \begin{aligned}
      P_u[0, 0] & \geq \cos(\theta) \\
      P_v[0, 0] & \geq \cos(\theta) \\
    \end{aligned}  
  \] 
  Therefore,  
  \begin{equation}\label{eq:error_esti}
      \begin{aligned}
          \|R_i - T_u \diag(\bar{S}) T_v\|_F^2 & =  \min_S \|\mathcal{U}\Sigma\mathcal{V} - \mathcal{U} P_u \diag(S) P_v \mathcal{V}\|^2_F\\
                                               & =  \min_S \|\Sigma - P_u \diag(S) P_v\|^2_F \\
      \end{aligned}
  \end{equation}
  By applying the definition of linear mapping $H: S \rightarrow \Sigma $ in eq \eqref{eq:H_define}, we can expand eq \eqref{eq:error_esti} further as follows:
  \begin{equation}\label{eq:error_esti_2}
      \begin{aligned}
          &  \|R_i - T_u \diag(\bar{S}) T_v\|_F^2 \\
          &= \|\Sigma\|_F^2 - \left\langle H^{\top}(\Sigma)  \left| (H^{\top}H)^{\dagger} \right| H^{\top}(\Sigma) \right\rangle \\
          &= \|\Sigma\|_F^2 \\
          &- \left\langle \underbrace{\diag(P_u^\top \Sigma P_v^\top)}_{\textnormal{scalar, cause of }q=1} \left|\underbrace{[(P_u^\top P_u) \odot (P_v P_v^\top)]^{\dagger}}_{=1}\right|\underbrace{\diag(P_u^\top \Sigma P_v^\top)}_{\textnormal{scalar, cause of }q=1} \right\rangle \\
          &= \|\sigma_i\|_2^2 - \left(\sum_k P_u[k, 0] \sigma_i[k] P_v[0, k]\right)^2
      \end{aligned}
  \end{equation}
  Noticing that:
  \begin{equation}\label{eq:error_esti_3}
      \begin{aligned}
          &\left|\sum_k P_u[k, 0] \sigma_i[k] P_v[0, k]\right| \\
          \geq & P_u[0, 0]\sigma_i[0]P_v[0, 0] - \left|\sum_{k\geq 1} P_u[k, 0] \sigma_i[k] P_v[0, k]\right| \\
          \geq & \sigma_i[0] \cos^2(\theta) - \underbrace{\sqrt{\sum_{k\geq 1} (P_u[k, 0])^2 \sigma_i[k]}\sqrt{\sum_{k\geq 1} (P_v[0, k])^2 \sigma_i[k]}}_{\textnormal{Cauchy inequality}} \\
          \geq & \sigma_i[0] \cos^2(\theta) - \sigma_i[0] \|P_u[1:, 0]\|_2 \|P_v[0, 1:]\|_2 \\
          = & \sigma_i[0] \cos^2(\theta) - \sigma_i[0] \sin^2(\theta) = \sigma_i[0] \cos(2\theta)
      \end{aligned}
  \end{equation}
  Now, by substituting eq \eqref{eq:error_esti_3} into eq \eqref{eq:error_esti_2}, we can obtain:
  \[
      \begin{aligned}
          & \frac{\|R_i - T_u \diag(\bar{S}) T_v\|_F^2}{\|R_i\|^2_F } \\
          \leq & \frac{\|\sigma_i\|^2_2 - (\sigma_i[0])^2 \cos^2(2\theta)}{\|R_i\|^2_F} \\
          =& \frac{\sin^2(2\theta)(\sigma_i[0])^2 + \|\sigma_i[1:]\|^2_2}{\|R_i\|^2_F } \\
          =& \frac{\sin^2(2\theta)(\sigma_i[0])^2 + \|\sigma_i[1:]\|^2_2}{(\sigma_i[0])^2 + \|\sigma_i[1:]\|^2_2} \\
          \leq & \frac{\sin^2(2\theta)(\sigma_i[0])^2 + (\min(M, N)-1)(\sigma_i[0])^2}{(\sigma_i[0])^2 + (\min(M, N)-1)(\sigma_i[0])^2} \\
          = & 1 - \frac{\cos^2(2\theta)}{\min(M, N)}   
      \end{aligned}
      \]
\end{proof}

\section{Experiment Detail}
\subsection{Experiment Configuration on Figure \ref{fig:TSVD_profile}}\label{sec:TSVD_profile}
In figure \ref{fig:TSVD_profile}, we utilize a self-consistent method to translate the computation costs among quantization, pruning, SVD, and our TSVD, as shown in Table \ref{tab:compute_cost}. The self-consistency can be verified by observing that the translation of TSVD is precisely equivalent to W$2$A$d$ quantization with sparsity consideration. For error, we select relative error of operator norm in figure \ref{fig:TSVD_profile}.

\subsection{Experiment Configuration on Figure \ref{fig:conv_select}}\label{sec:conv_select}

As the unfold matrix $\mathcal{W}$ is sparse, and it is important that \emph{the FLOPS of $\mathcal{W}(\cdot)$ is only equal to its folding version when considering the sparsity of $\mathcal{W}$}. Therefore, in figure \ref{fig:conv_select}, we have modified translation method of TSVD in table \ref{tab:compute_cost} as:
\[
\text{compress rate (TSVD)} = \frac{K(d-2) + rK(M+N)}{r'MN(d-1)}
\]
where $r'$ is the sparsity rate of $\mathcal{W}$. The compression rate of Winograd $F(2\times 2, 3\times 3)$ is also calculated in this way. All data points in figure \ref{fig:conv_select} are the mean of three runs.

\subsection{Final Network Architecture in Section \ref{sec:imagenet_1k}} \label{sec:imagenet_1k_detail}
For the sake of brevity, we only present the final architecture of the 7\% tolerance TSVD with finetune of ConvNeXt-T model, and have omitted all layer information except for the TSVD layer. The architecture is displayed as below:

{\tiny\onecolumn\begin{verbatim}
(convnext): ConvNextModel(
  (embeddings): ConvNextEmbeddings(
    (patch_embeddings): Ternary_SVD_Conv2d(3, 96, kernel_size=(4, 4), stride=(4, 4), form_type=3, rank=63, sparsity=0.269)
  )
  (encoder): ConvNextEncoder(
    (stages): ModuleList(
      (0): ConvNextStage(
        (downsampling_layer): Identity()
        (layers): Sequential(
          (0): ConvNextLayer(
            (dwconv): Ternary_SVD_Conv2d(96, 96, kernel_size=(7, 7), stride=(1, 1), padding=(3, 3), groups=96, form_type=0,
              rank=5, sparsity=0.235)
            (pwconv1): Ternary_SVD_Linear(in_features=96, out_features=384, bias=True, rank=345, sparsity=0.277)
            (pwconv2): Ternary_SVD_Linear(in_features=384, out_features=96, bias=True, rank=384, sparsity=0.279)
          )
          (1): ConvNextLayer(
            (dwconv): Ternary_SVD_Conv2d(96, 96, kernel_size=(7, 7), stride=(1, 1), padding=(3, 3), groups=96, form_type=0,
              rank=5, sparsity=0.246)
            (pwconv1): Ternary_SVD_Linear(in_features=96, out_features=384, bias=True, rank=447, sparsity=0.277)
            (pwconv2): Ternary_SVD_Linear(in_features=384, out_features=96, bias=True, rank=414, sparsity=0.277)
          )
          (2): ConvNextLayer(
            (dwconv): Ternary_SVD_Conv2d(96, 96, kernel_size=(7, 7), stride=(1, 1), padding=(3, 3), groups=96, form_type=0,
              rank=5, sparsity=0.201)
            (pwconv1): Ternary_SVD_Linear(in_features=96, out_features=384, bias=True, rank=435, sparsity=0.278)
            (pwconv2): Ternary_SVD_Linear(in_features=384, out_features=96, bias=True, rank=417, sparsity=0.277)
          )
        )
      )
      (1): ConvNextStage(
        (downsampling_layer): Sequential(
          (1): Ternary_SVD_Conv2d(96, 192, kernel_size=(2, 2), stride=(2, 2), form_type=2, rank=648, sparsity=0.278)
        )
        (layers): Sequential(
          (0): ConvNextLayer(
            (dwconv): Ternary_SVD_Conv2d(192, 192, kernel_size=(7, 7), stride=(1, 1), padding=(3, 3), groups=192, form_type=0,
              rank=5, sparsity=0.283)
            (pwconv1): Ternary_SVD_Linear(in_features=192, out_features=768, bias=True, rank=763, sparsity=0.276)
            (pwconv2): Ternary_SVD_Linear(in_features=768, out_features=192, bias=True, rank=812, sparsity=0.276)
          )
          (1): ConvNextLayer(
            (dwconv): Ternary_SVD_Conv2d(192, 192, kernel_size=(7, 7), stride=(1, 1), padding=(3, 3), groups=192, form_type=0,
              rank=5, sparsity=0.246)
            (pwconv1): Ternary_SVD_Linear(in_features=192, out_features=768, bias=True, rank=819, sparsity=0.275)
            (pwconv2): Ternary_SVD_Linear(in_features=768, out_features=192, bias=True, rank=763, sparsity=0.277)
          )
          (2): ConvNextLayer(
            (dwconv): Ternary_SVD_Conv2d(192, 192, kernel_size=(7, 7), stride=(1, 1), padding=(3, 3), groups=192, form_type=0,
              rank=5, sparsity=0.231)
            (pwconv1): Ternary_SVD_Linear(in_features=192, out_features=768, bias=True, rank=819, sparsity=0.275)
            (pwconv2): Ternary_SVD_Linear(in_features=768, out_features=192, bias=True, rank=882, sparsity=0.275)
          )
        )
      )
      (2): ConvNextStage(
        (downsampling_layer): Sequential(
          (1): Ternary_SVD_Conv2d(192, 384, kernel_size=(2, 2), stride=(2, 2), form_type=2, rank=1152, sparsity=0.273)
        )
        (layers): Sequential(
          (0): ConvNextLayer(
            (dwconv): Ternary_SVD_Conv2d(384, 384, kernel_size=(7, 7), stride=(1, 1), padding=(3, 3), groups=384, form_type=0,
              rank=5, sparsity=0.280)
            (pwconv1): Ternary_SVD_Linear(in_features=384, out_features=1536, bias=True, rank=1545, sparsity=0.275)
            (pwconv2): Ternary_SVD_Linear(in_features=1536, out_features=384, bias=True, rank=1380, sparsity=0.275)
          )
          (1): ConvNextLayer(
            (dwconv): Ternary_SVD_Conv2d(384, 384, kernel_size=(7, 7), stride=(1, 1), padding=(3, 3), groups=384, form_type=0,
              rank=5, sparsity=0.231)
            (pwconv1): Ternary_SVD_Linear(in_features=384, out_features=1536, bias=True, rank=1530, sparsity=0.275)
            (pwconv2): Ternary_SVD_Linear(in_features=1536, out_features=384, bias=True, rank=1470, sparsity=0.275)
          )
          (2): ConvNextLayer(
            (dwconv): Ternary_SVD_Conv2d(384, 384, kernel_size=(7, 7), stride=(1, 1), padding=(3, 3), groups=384, form_type=0,
              rank=5, sparsity=0.231)
            (pwconv1): Ternary_SVD_Linear(in_features=384, out_features=1536, bias=True, rank=1485, sparsity=0.275)
            (pwconv2): Ternary_SVD_Linear(in_features=1536, out_features=384, bias=True, rank=1560, sparsity=0.275)
          )
          (3): ConvNextLayer(
            (dwconv): Ternary_SVD_Conv2d(384, 384, kernel_size=(7, 7), stride=(1, 1), padding=(3, 3), groups=384, form_type=0,
              rank=5, sparsity=0.241)
            (pwconv1): Ternary_SVD_Linear(in_features=384, out_features=1536, bias=True, rank=1575, sparsity=0.275)
            (pwconv2): Ternary_SVD_Linear(in_features=1536, out_features=384, bias=True, rank=1545, sparsity=0.276)
          )
          (4): ConvNextLayer(
            (dwconv): Ternary_SVD_Conv2d(384, 384, kernel_size=(7, 7), stride=(1, 1), padding=(3, 3), groups=384, form_type=0,
              rank=5, sparsity=0.256)
            (pwconv1): Ternary_SVD_Linear(in_features=384, out_features=1536, bias=True, rank=1500, sparsity=0.275)
            (pwconv2): Ternary_SVD_Linear(in_features=1536, out_features=384, bias=True, rank=1650, sparsity=0.276)
          )
          (5): ConvNextLayer(
            (dwconv): Ternary_SVD_Conv2d(384, 384, kernel_size=(7, 7), stride=(1, 1), padding=(3, 3), groups=384, form_type=0,
              rank=5, sparsity=0.280)
            (pwconv1): Ternary_SVD_Linear(in_features=384, out_features=1536, bias=True, rank=1515, sparsity=0.275)
            (pwconv2): Ternary_SVD_Linear(in_features=1536, out_features=384, bias=True, rank=1605, sparsity=0.275)
          )
          (6): ConvNextLayer(
            (dwconv): Ternary_SVD_Conv2d(384, 384, kernel_size=(7, 7), stride=(1, 1), padding=(3, 3), groups=384, form_type=0,
              rank=5, sparsity=0.287)
            (pwconv1): Ternary_SVD_Linear(in_features=384, out_features=1536, bias=True, rank=1440, sparsity=0.275)
            (pwconv2): Ternary_SVD_Linear(in_features=1536, out_features=384, bias=True, rank=1560, sparsity=0.276)
          )
          (7): ConvNextLayer(
            (dwconv): Ternary_SVD_Conv2d(384, 384, kernel_size=(7, 7), stride=(1, 1), padding=(3, 3), groups=384, form_type=0,
              rank=5, sparsity=0.251)
            (pwconv1): Ternary_SVD_Linear(in_features=384, out_features=1536, bias=True, rank=1365, sparsity=0.275)
            (pwconv2): Ternary_SVD_Linear(in_features=1536, out_features=384, bias=True, rank=1590, sparsity=0.276)
          )
          (8): ConvNextLayer(
            (dwconv): Ternary_SVD_Conv2d(384, 384, kernel_size=(7, 7), stride=(1, 1), padding=(3, 3), groups=384, form_type=0,
              rank=5, sparsity=0.252)
            (pwconv1): Ternary_SVD_Linear(in_features=384, out_features=1536, bias=True, rank=1350, sparsity=0.276)
            (pwconv2): Ternary_SVD_Linear(in_features=1536, out_features=384, bias=True, rank=1785, sparsity=0.275)
          )
        )
      )
      (3): ConvNextStage(
        (downsampling_layer): Sequential(
          (1): Ternary_SVD_Conv2d(384, 768, kernel_size=(2, 2), stride=(2, 2), form_type=2, rank=1975, sparsity=0.275)
        )
        (layers): Sequential(
          (0): ConvNextLayer(
            (dwconv): Ternary_SVD_Conv2d(768, 768, kernel_size=(7, 7), stride=(1, 1), padding=(3, 3), groups=768, form_type=0,
              rank=5, sparsity=0.283)
            (pwconv1): Ternary_SVD_Linear(in_features=768, out_features=3072, bias=True, rank=1740, sparsity=0.275)
            (pwconv2): Ternary_SVD_Linear(in_features=3072, out_features=768, bias=True, rank=3570, sparsity=0.275)
          )
          (1): ConvNextLayer(
            (dwconv): Ternary_SVD_Conv2d(768, 768, kernel_size=(7, 7), stride=(1, 1), padding=(3, 3), groups=768, form_type=0,
              rank=5, sparsity=0.273)
            (pwconv1): Ternary_SVD_Linear(in_features=768, out_features=3072, bias=True, rank=1560, sparsity=0.275)
            (pwconv2): Ternary_SVD_Linear(in_features=3072, out_features=768, bias=True, rank=3390, sparsity=0.275)
          )
          (2): ConvNextLayer(
            (dwconv): Ternary_SVD_Conv2d(768, 768, kernel_size=(7, 7), stride=(1, 1), padding=(3, 3), groups=768, form_type=0,
              rank=5, sparsity=0.281)
            (pwconv1): Ternary_SVD_Linear(in_features=768, out_features=3072, bias=True, rank=1530, sparsity=0.275)
            (pwconv2): Ternary_SVD_Linear(in_features=3072, out_features=768, bias=True, rank=3840, sparsity=0.275)
          )
        )
      )
    )
  )
)
(classifier): Ternary_SVD_Linear(in_features=768, out_features=1000, bias=True, rank=2415, sparsity=0.275)
\end{verbatim}}
\end{document}